\documentclass[letterpaper,10pt,conference]{ieeeconf}
\IEEEoverridecommandlockouts
\let\labelindent\relax
\usepackage[utf8]{inputenc} %
\usepackage[T1]{fontenc}    %
\usepackage{graphicx} \graphicspath{ {figures/} }
\usepackage{amsmath,amssymb,mathabx,amsbsy,mathtools,etoolbox}
\usepackage{times}
\usepackage{graphicx}
\usepackage{amsmath,amssymb,mathbbol}
\usepackage{acronym}
\usepackage{enumitem}
\usepackage[breaklinks,colorlinks,linkcolor=black]{hyperref}
\usepackage{balance}
\usepackage{xspace}
\usepackage{setspace}
\usepackage[skip=3pt,font=small]{subcaption}
\usepackage[skip=3pt,font=small]{caption}
\usepackage[dvipsnames,svgnames,x11names]{xcolor}
\usepackage[capitalise,noabbrev,nameinlink]{cleveref}
\usepackage{booktabs,tabularx,colortbl,multirow,multicol,array,makecell}
\usepackage{dblfloatfix}
\usepackage[misc]{ifsym}
\usepackage{pifont}
\usepackage{cite}
\usepackage{graphicx}

\makeatletter
\DeclareRobustCommand\onedot{\futurelet\@let@token\@onedot}
\def\@onedot{\ifx\@let@token.\else.\null\fi\xspace}

\def\etal{\emph{et al}\onedot}
\makeatother

\DeclareMathOperator*{\argmin}{argmin}

\makeatletter
\def\BState{\State\hskip-\ALG@thistlm}
\makeatother

\makeatletter
\renewcommand{\paragraph}{%
  \@startsection{paragraph}{4}%
  {\z@}{0ex \@plus 0ex \@minus 0ex}{-1em}%
  {\hskip\parindent\normalfont\normalsize\bfseries}%
}
\makeatother

\crefname{section}{Sec.}{Secs.}
\Crefname{section}{Section}{Sections}
\crefname{table}{Tab.}{Tabs.}
\Crefname{table}{Table}{Tables}
\crefname{figure}{Fig.}{Figs.}
\Crefname{figure}{Figure}{Figure}
\Crefname{equation}{Eq.}{Eq.}

\usepackage{pifont}
\usepackage{bbm}
\usepackage{soul}
\usepackage[export]{adjustbox}

\newcommand{\LL}{\mathcal{L}}

\newcommand{\HH}{\mathbf{H}}
\newcommand{\OO}{\mathbf{O}}
\newcommand{\TT}{\mathbf{T}}
\newcommand{\FF}{\mathbf{F}}

\title{GrainGrasp: Dexterous Grasp Generation with Fine-grained Contact Guidance}
\author{Fuqiang Zhao$^{1}$, Dzmitry Tsetserukou$^{2}$ and Qian Liu$^{1 \ast}$
\thanks{
This work was supported in part by the National Science Foundation of China(Grant No.62071083), and in part by the Dalian Science and Technology Innovation Foundation (No. 2022JJ12GX014).}
\thanks{${^1}$Fuqiang Zhao and Qian Liu are with the the Department of Computer Science and Technology, Dalian University of Technology, China. Emails: fuqiangzh@mail.dlut.edu.cn, qianliu@dlut.edu.cn.}
\thanks{${^2}$Dzmitry Tsetserukou is with the Skolkovo Institute of Science and Technology, 121205 Moscow, Russia. Email: D.Tsetserukou@skoltech.ru.}
\thanks{$^{\ast}$Corresponding author: Qian Liu.}
}

\begin{document}

\maketitle
\thispagestyle{empty}

\begin{abstract}
    One goal of dexterous robotic grasping is to allow robots to handle objects with the same level of flexibility and adaptability as humans. However, it remains a challenging task to generate an optimal grasping strategy for dexterous hands, especially when it comes to delicate manipulation and accurate adjustment the desired grasping poses for objects of varying shapes and sizes. In this paper, we propose a novel dexterous grasp generation scheme called \textbf{\textit{GrainGrasp}} that provides fine-grained contact guidance for each fingertip. In particular, we employ a generative model to predict separate contact maps for each fingertip on the object point cloud, effectively capturing the specifics of finger-object interactions. In addition, we develop a new dexterous grasping optimization algorithm that solely relies on the point cloud as input, eliminating the necessity for complete mesh information of the object. By leveraging the contact maps of different fingertips, the proposed optimization algorithm can generate precise and determinable strategies for human-like object grasping. Experimental results confirm the efficiency of the proposed scheme. Our code is available at~\href{https://github.com/wmtlab/GrainGrasp}{\textit{https://github.com/wmtlab/GrainGrasp}}.
\end{abstract}
\setstretch{0.95}

\section{Introduction}
The past decades have witnessed the rapid development of high-fidelity humanoid hands in computer graphics, e.g. MANO~\cite{mano}, as well as real-world dexterous robotic hands, e.g. Shadow Hand~\cite{shadowrobot}. These advancements have enabled realistic hand pose imitation and dexterous manipulation, with widespread applications in Virtual Reality\!~\cite{han2022umetrack},\!\cite{holl2018efficient}, Human-Computer Interactions~\cite{xue2020progress} and \nocite{breyer2021volumetric,ye2023learning,qin2022one,thumbcontact2019contactgrasp,li2023gendexgrasp}robotics~\cite{breyer2021volumetric}-\!\!\cite{li2023gendexgrasp}.

\begin{figure}
\centering
\includegraphics[width=0.49\textwidth]{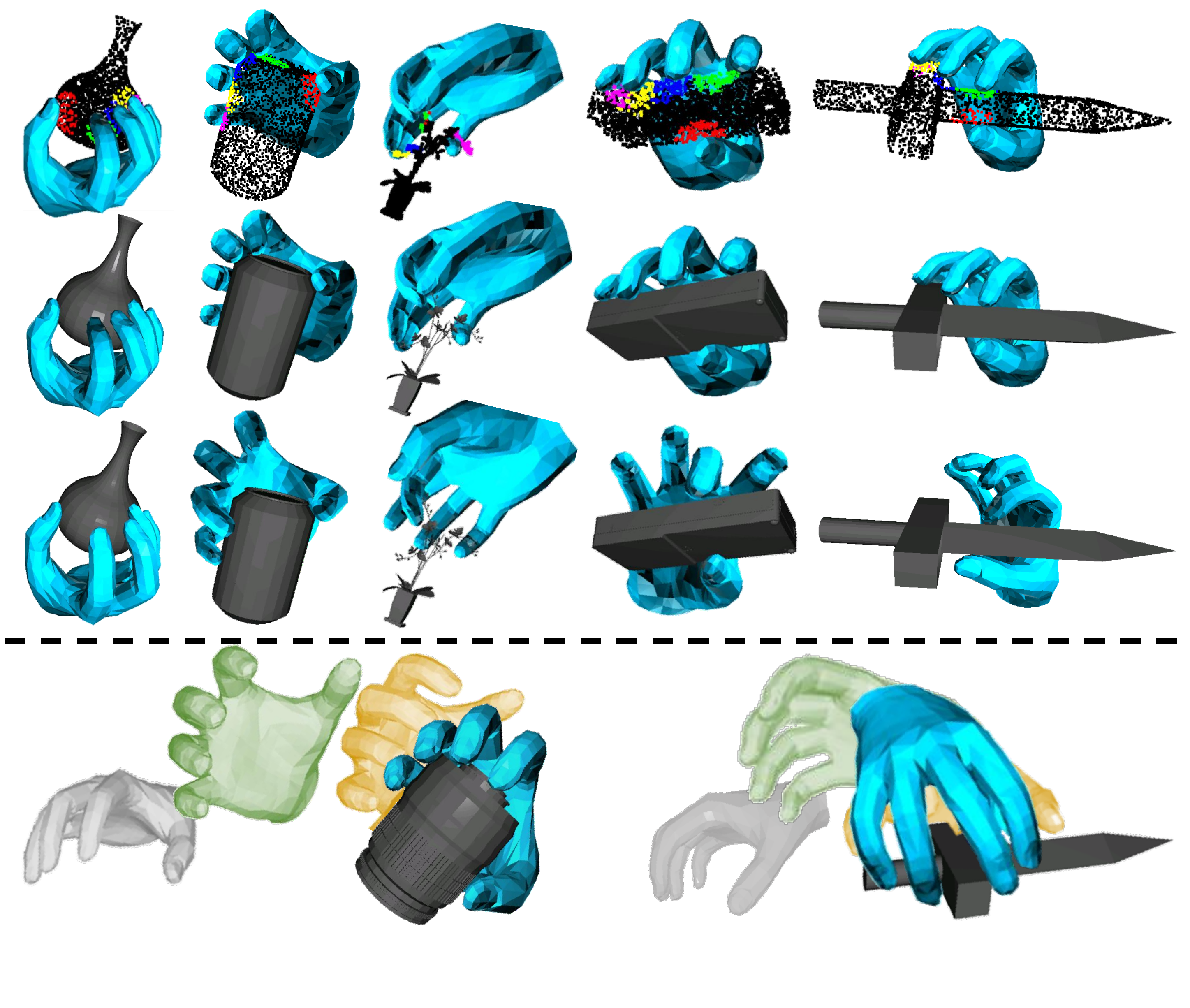} \vspace{-1cm}
\caption{Examples of grasping results and grasping processes generated by the GrainGrasp. Top: The grasp results are displayed with point cloud and mesh representations in the first two rows. The colored points in the point cloud indicate the contact map of each finger. The third row illustrates the determinable results achieved by controlling the contribution of each finger. Bottom: The grasping process is represented in four steps: gray $\rightarrow$ green $\rightarrow$ yellow $\rightarrow$ blue. Throughout the process, the hand first adjusts its direction, then quickly approaches the object, which presents human-like characteristics of object grasping.}
\label{fig:example}
\end{figure}

The advances of digital and robotic hands have promoted dexterous grasping as an increasingly important research direction. Recent studies focus on obtaining diverse and high-quality grasps, including data-driven approaches, optimization algorithms, as well as combinations of these techniques. A typical procedure is to first utilize a Deep Learning algorithm to predict the contact maps from objects in datasets, then employ an optimization algorithm or a reinforcement learning approach to guide the hand to match the predicted contact maps and achieve grasping. Unfortunately, we should admit that the predicted the contact maps of this typical procedure can only provide a coarse approximation of ideal contacts. The ambiguity in the contact map makes it difficult to achieve stable grasps for complex objects. In order to achieve better grasping, it is necessary to adjust the position and force of each individual finger on the dexterous hand with fine-grained contact maps. However, existing grasp generation methods generally produce grasps without detailed finger adjustments, resulting in unsatisfactory performance on complex tasks. Therefore, realizing fine granularity over each finger becomes an essential task for high-quality grasp generation. 

On the other hand, due to the lack of explicit modeling between contact maps and executed grasps, even with fine-grained contact maps, it is not guaranteed to achieve human-desired grasping positions. This limitation poses a challenge to the effectiveness of using contact map prediction as a reliable supervision signal. Although methods like ContactGrasp~\cite{thumbcontact2019contactgrasp} set specific contact locations for the thumb, controlling the thumb alone is insufficient for dexterous manipulation, which requires coordinated motion of multiple fingers. The key to resolving this problem lies in establishing a direct mapping and conducting joint optimization of finger positions, incorporating appropriate contact constraints.

\begin{figure*}[t!]
    \centering
    \includegraphics[width=\linewidth]{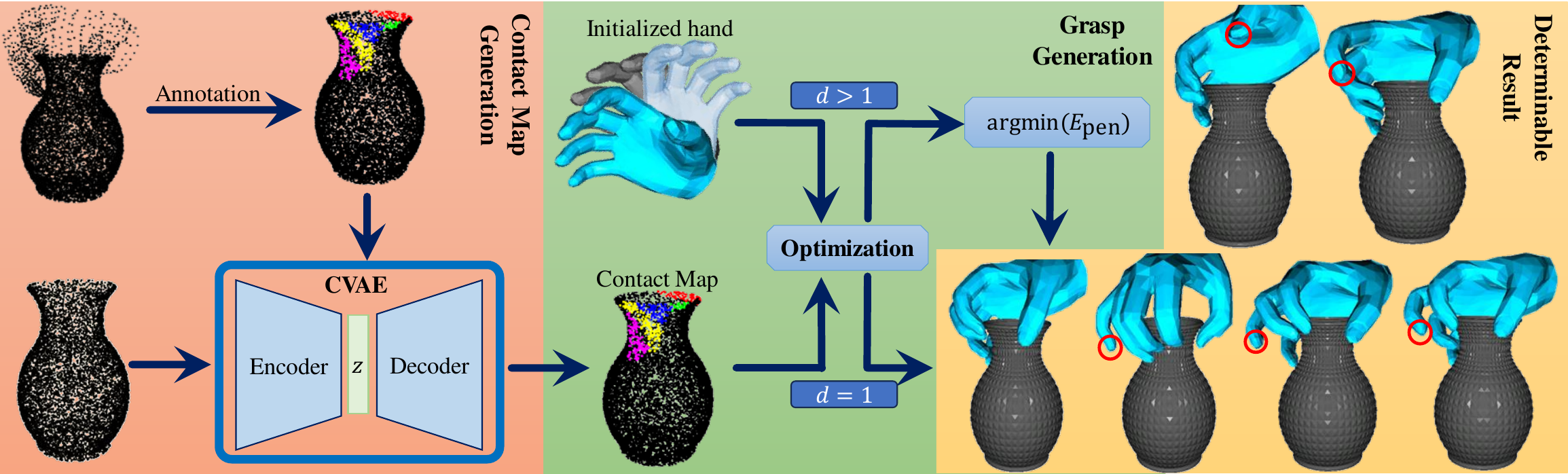}
    \caption{Pipeline of our method. \textbf{1)} We automatically annotate point cloud data and train a CVAE model to generate individual contact maps. \textbf{2)} We utilize the object point cloud and contact maps to optimize the initialized hand pose. If the number of directions $d$ rotates more than once (i.e., $d > 1$), the final grasping result is obtained by minimizing $E_\text{pen}$ under the condition that $E_\text{pen} > 0$. \textbf{3)} We generate determinable grasping results by adjusting individual finger contributions.}
    \label{fig:pipeline}
\end{figure*} 

To address the above-mentioned limitations, we reformulate the whole-hand grasping task by focusing on managing the individual contribution and coordination of each finger, rather than regulating the entire hand. In order to obtain a fine-grained contact map for grasping, we propose to predict a distinct contact map for each fingertip. This allows us to capture the details of finger-object contacts, instead of predicting a holistic contact map or only a small number of contact points as that of existing methods. Inspired by previous work~\cite{li2023gendexgrasp},\!\cite{jiang2021hand}, we adopt generative models for this research.

Furthermore, we propose the \textbf{DCoG} (Directional Consistency optimization for Grasping) to comprehensively model the intricate relationship between the contact map and the grasping determination. DCoG emphasizes the consistency in the physically appropriate direction of each finger between the finger and the object surface during the grasping process. This way, DCoG can reveal the role of each finger in the grasping task and its coordinated relationship with other fingers. As a result, it provides a fine-grained guidance to the grasping behavior of each finger, enabling advanced grasping generation. Examples of grasping results and the corresponding grasping processes generated by the proposed scheme can be found in~\cref{fig:example}.

In summary, the proposed scheme consists of two key components: the Contact Map Generation and the Grasp Generation modules, as shown in~\cref{fig:pipeline}. We further summarize the contributions of this research as follows.
\begin{enumerate}[leftmargin=*,nolistsep,noitemsep,topsep=-1em,labelsep=0.1em,label=\textbf{\arabic*}),]
\item We propose a new method to predict distinct contact maps for individual fingertips. This method offers enhanced precision in achieving a stable grasp by individually adjusting the contact of each finger, rather than making direct adjustments to the entire hand. Consequently, it can provide superior guidance for achieving grasping that closely resembles human capabilities.
\item Our proposed optimization method, known as DCoG, allows for interpretable and determinable robotic grasping using contact maps. This contact-based grasping algorithm enables explainable and intuitive robot manipulation.
\end{enumerate}

\section{Related Work}

    \subsection{Dexterous Grasp Synthesis}

    The synthesis of dexterous grasps poses significant challenges due to the numerous degrees of freedom and complexity of modeling grasp interactions. 
    
    Analytical methods typically focus on optimizing the grasp pose to achieve force closure with consideration of physical constraints\!\!\nocite{li2003computing, sahbani2012overview, rosales2012synthesis, prattichizzo2012manipulability}
    \!\!~\cite{li2003computing}-\!\!\!\cite{prattichizzo2012manipulability}. Recently, Liu~\etal\cite{liu2021dfc} developed a differentiable force closure estimator that enables the generation of varied and physically stable grasps for robotic hands with arbitrary structures. Building upon this method, Wang \etal~\cite{wang2023dexgraspnet} refined the initial hand pose, and contributed a new grasp dataset called DexGraspNet. Similarly, Li \etal~\cite{li2023gendexgrasp} collected a dataset of grasps from different hands using the method described in \cite{liu2021dfc}.
    
    With the development of Deep Learning and grasp datasets
    \nocite{datasetchao2021dexycb, datasethampali2020honnotate, datasethasson2019learning, brahmbhatt2019contactdb, brahmbhatt2020contactpose, wang2023dexgraspnet}
    ~\cite{wang2023dexgraspnet}-\!\!\!\cite{brahmbhatt2020contactpose}, data-driven methods have gained immense popularity. These methods utilize Deep Neural Networks to learn from datasets, enabling direct predictions of grasping parameters or critical information for grasping, such as contact points and contact maps. GraspCVAE~\cite{jiang2021hand} and GraspGlow~\cite{xu2023unidexgrasp} generated corresponding grasping parameters for 3D objects. These methods generally employed test-time adaptation (TTA) to refine the generated poses. UniGrasp~\cite{shao2020unigrasp} and EfficientGrasp~\cite{li2022efficientgrasp} utilized a neural network called PSSN to select contact points. ContactNet~\cite{jiang2021hand},\cite{xu2023unidexgrasp} and DeepContact~\cite{grady2021contactopt} were trained to estimate contact maps. Additionally, other data-driven methods also showed promising results. One example is the training of Grasping Field~\cite{karunratanakul2020graspingfield} as an implicit function which used the signed distances~\cite{park2019deepsdf} of 3D points to infer contact regions. Another example is CGF~\cite{ye2023learning}, which trained a Conditional Variational Auto-Encoder (CVAE)~\cite{cvae} framework to generate continuous trajectories.

    \subsection{Contact-based Approaches}

    In robotic grasping, contact information visually represents the physical interactions between the robot hand and the target object. Numerous research studies utilize this information to enhance grasping capabilities.

    UniGrasp~\cite{shao2020unigrasp} predicted contact points on the object and adjusted the gripper trajectory using inverse kinematics. Wu \etal~\cite{wu2022learning} employed the CVAE to predict finger placements, which were used to initialize a Bilevel Optimization for grasp configuration. ContactOpt~\cite{grady2021contactopt} trained the DeepContact model by performing multiple random perturbations on hand-object grasps using the ContactPose dataset~\cite{brahmbhatt2020contactpose} to generate the target contact map. S$^2$Contact~\cite{s2contact2022s} utilized visual and geometric consistency constraints to generate pseudo-labels for semi-supervised learning and employed a graph-based network to infer the contact map. ContactGrasp~\cite{thumbcontact2019contactgrasp} sampled random grasps using GraspIt!~\cite{miller2004graspit} and ranked them based on their consistency with a thumb-aligned contact map to synthesize functional grasps using DART~\cite{schmidt2014dart}. Jiang \etal~\cite{jiang2021hand} proposed a self-supervised approach to improve grasping. It updated the hand pose based on the difference between the contact map predicted by ContactNet and the contact map estimated by the distance between the hand and the object during inference. Similar to the approach presented in~\cite{jiang2021hand}, UniDexGrasp~\cite{xu2023unidexgrasp} incorporates information from the predicted contact map generated by ContactNet. This information is used to optimize the grasping pose during the intermediate step of the grasping process. GenDexGrasp~\cite{li2023gendexgrasp} trained a generative model CVAE to generate a contact map for the object point cloud, then used an optimization algorithm to fit the hand to the generated contact map. Mandikal \etal~\cite{mandikal2021learning} modeled the contact map as the object's affordance and treated the affordance learning problem as a segmentation task to predict binary per-pixel labels on the contact map.
    
    Inspired by the aforementioned methods, we approach the generation of contact maps as a point cloud segmentation task. In this study, we generate individual contact maps for each fingertip on the object point cloud. By predicting contact maps per fingertip, we can independently optimize the contact location and force for each finger. Consequently, our proposed approach offers fine-grained grasping guidance for a dexterous hand.

\begin{figure}
    \centering
    \begin{subfigure}{0.32\linewidth}
        \centering
        \includegraphics[width=\linewidth,height=0.95\linewidth]{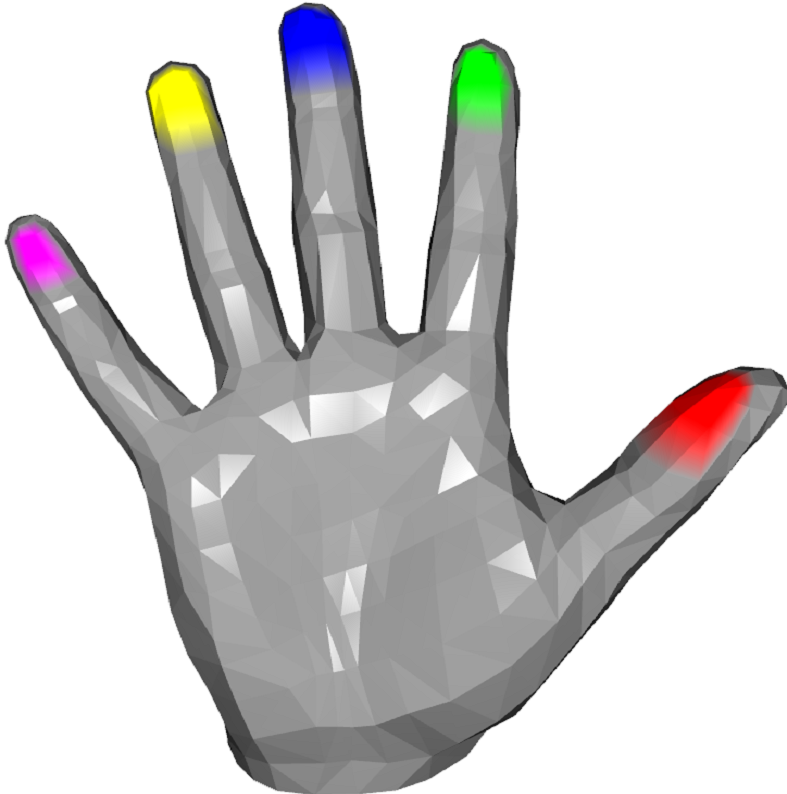}
        \caption{}
        \label{fig:annotation:fingertip}
    \end{subfigure}
    \centering
    \begin{subfigure}{0.32\linewidth}
        \centering
        \includegraphics[width=\linewidth,height=0.95\linewidth]{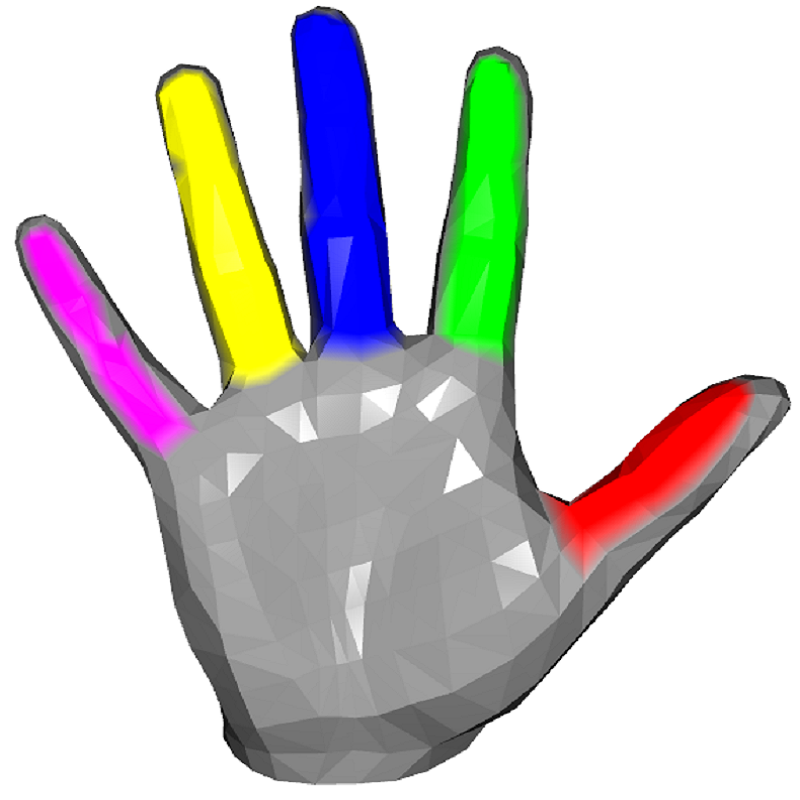}
        \caption{}
        \label{fig:annotation:finger}
    \end{subfigure}
    \centering
    \begin{subfigure}{0.32\linewidth}
        \centering
        \includegraphics[width=\linewidth,height=0.95\linewidth]{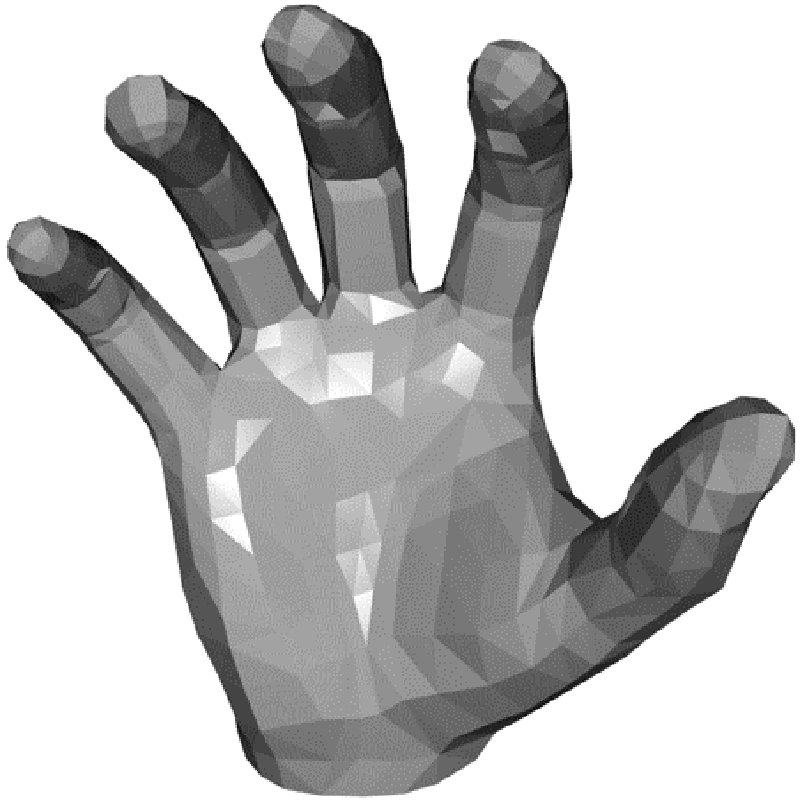}
        \caption{}
        \label{fig:annotation:inithand}
    \end{subfigure}
    
    \caption{(a)\&(b) Five fingertips and fingers are presented in different colors for distinction. (c) Our initial hand pose exhibits a slight flexion at the interphalangeal joint compared to a fully flat hand.}
    \label{fig:annotation}
\end{figure}

\section{Contact Map Generation}
    The development of an efficient separate contact map generator involves a two-step process. We first automatically annotate the ObMan dataset~\cite{datasethasson2019learning} and then use it to train a CVAE model. This CVAE model can generate separate contact maps for each finger when given an input of object point cloud.

    \subsection{Dataset Annotation}     \label{sec:annotate}
    
    Unlike previous algorithms~\cite{thumbcontact2019contactgrasp},\!\cite{grady2021contactopt} that directly utilize thermal contacts from the ContactDB~\cite{brahmbhatt2019contactdb} and ContactPose~\cite{brahmbhatt2020contactpose} datasets, our method enables automatic annotation of contact maps on the Grasp dataset solely relies on hand-object point cloud data. In this research, we adopt the ObMan dataset~\cite{datasethasson2019learning} and conduct all experiments based on the MANO hand model, which can be described by $\theta_H=(\theta,t,R)$, with $\theta, t, R$ indicating the pose, translation, and rotation, respectively. The pose $\theta$ is represented as a 45-dimensional Principal Component Analysis (PCA) manifold. It should be noted that the MANO model also requires shape parameters to describe the hand model. In this paper, we use the default shape parameter, which remains unchanged during the optimization process.
    
    \textit{First}, we manually delineate the fingertip region of the hand using the Blender software~\cite{blender}. As shown in Fig. 3(a), the five fingertips are delineated in different colors, indicating a specific fingertip encountering the object on the annotated contact map. \textit{Second}, for points within the delineated regions representing the five fingertips, we compute the K nearest points on the object point cloud. An issue may arise during this step, where some points may belong to contact maps of multiple fingertips simultaneously. To resolve this problem, we should determine which fingertip region is closest to each point and assign it accordingly in the contact map. \textit{Finally}, we classify each point in the object point cloud into six categories: contacted by the thumb, index, middle, ring, pinky fingertip, or uncontacted, respectively.

    Since we only use object point clouds, we can only rely on Euclidean distance instead of the aligned distance adopted by Li \textit{et al}.~\cite{li2023gendexgrasp}. Consequently, some false contact maps may be generated when annotating thin objects. Therefore, we cannot have a large K. On the other hand, a smaller K value can lead to a reduction in the overall size of annotated contact maps, which can cause difficulty in model prediction. Therefore, we encourage the utilization of a more appropriate K value. Based on our experiments, it seems that setting the value of K to 50 for a point cloud of 3000 points resulted in satisfactory point labeling performance.

    For clearer illustration, we denote the subset involving the five contacted categories as $C$. Then, we employ the notations $\TT$, $\HH$, and $\OO$ to symbolize the point cloud sets corresponding to fingertips, full hands, and objects, respectively. We concurrently manually delineated the finger region, as shown in Fig. 3(b). We denote these labeled point cloud sets as $\FF$, which will be used in the proposed optimization algorithm in~\cref{sec:optim}.

\begin{table*}[tb]
    \setlength{\abovecaptionskip}{0cm}
    \setlength{\belowcaptionskip}{0.2cm}
    \centering
    \caption{\label{table1} Comparison Experiments} 
    \begin{tabular}{lccccc}
        \toprule
        \textbf{Methods} & \textbf{Volume$(cm^3)\downarrow$} & \textbf{Depth$(cm)\downarrow$} & \textbf{Displacement$(cm)\downarrow$} & \textbf{Succ. Rate$(\mathrm{\%.})\uparrow$} & \textbf{Perc. Scores$\uparrow$} \\ \midrule
        GT~\cite{datasethasson2019learning} & 2.20 & 0.66 & \textbf{0.75} & \textbf{52.28} & 3.51 \\ \midrule
        Contactopt~\cite{grady2021contactopt} & 3.25 & 0.68 & 0.91 & 18.52  & - \\ \midrule
        GF~\cite{karunratanakul2020graspingfield} & 2.38 & 0.94 & 0.89 & 20.60 & 2.93 \\ \midrule
        GA (w/o TTA)~\cite{jiang2021hand} & 2.99 & 0.77 & 0.87 & 33.78 & - \\ 
        GA (w/ TTA)~\cite{jiang2021hand} & 3.65 & 0.75 & \textbf{0.80} & 35.61 & 3.43\\ \midrule
        Ours (only opt.) & \textbf{1.48} & 0.57 & 0.82 & \textbf{45.51} & \textbf{3.66} \\
        Ours (complete) & 1.79 & \textbf{0.48} & 0.84 & 43.10 & 3.41 \\ \bottomrule
    \end{tabular}%

    \vspace{0.1cm} 

    \begin{minipage}{\linewidth}
        \centering
        \footnotesize
        GT represents results from the dataset. GF represents results from the Grasping Field method. GA represents results from the GraspTTA method.
    \end{minipage}
\end{table*}

    \subsection{CVAE Synopsis}
    \label{sec:cvae}
    Inspired by~\cite{li2023gendexgrasp},\!~\cite{shao2020unigrasp}, we regard the contact map generating problem as a segmentation task to predict multiclass pointwise labels. Meanwhile, we adopt the CVAE~\cite{cvae} to generate the separate contact maps of fingertips. 
 
    CVAE consists of an encoding stage and a decoding stage. For the \textbf{encoding} stage, we first utilize the PointNet encoder~\cite{qi2017pointnet} to extract pointwise features from the object point cloud. In parallel, an embedding layer is employed to obtain embedded category features from the corresponding contact map. Then, we concatenate the two features and apply Multi-Layer Perceptrons (MLP), followed by a max pooling layer, to obtain the latent distribution $\mathcal{N}(\mu,\sigma^2)$. For the \textbf{decoding} stage, we sample the latent code $z\sim\mathcal{N}(\mu,\sigma^2)$ and replicate it $n$ times, where $n$ indicates the number of points in the point cloud. The replicated latent code is concatenated with the pointwise features and passed through an MLP to generate the contact map.

    For the training of the CVAE model, we leverage two types of supervised loss functions: a reconstruction loss that penalizes errors in reconstructing the contact map and a regularization loss that encourages the latent space to have desirable properties.

    The reconstruction loss consists of a multiclass cross-entropy loss $\LL_{\text{cross}}$ 
    and a multiclass dice loss $\LL_{\text{dice}}$~\cite{milletari2016dice},
    which enables the network to generate more accurate contact map from the input point cloud.
    We use a KL loss $\LL_{\text{KD}}$ as the regularization loss. The $\LL_{\text{KD}}$ is defined by maximizing the KL-Divergence between the latent code distribution $\mathcal{N}(\mu,\sigma^2)$ and a standard Gaussian distribution $\mathcal{N}(0,1)$, which encourages the latent code distribution learned by the encoder to be close to the prior standard Gaussian distribution. The complete loss function to train the CVAE can be expressed as:
    \begin{equation}
    \LL = \lambda_{\text{cross}}\LL_{\text{cross}} +  \lambda_{\text{dice}}\LL_{\text{dice}} + \lambda_{\text{KD}}\LL_{\text{KD}}
    \end{equation}
    where $\lambda_{\text{cross}}$, $\lambda_{\text{dice}}$ and $\lambda_{\text{KD}}$ represent the weights of corresponding losses. Their values are set as $0.5$, $0.9$ and $0.01$, respectively.

\section{Proposed Grasping Optimization}

    When a human grasps an object, the fingertips apply force in the direction perpendicular to the surface to ensure a stable grip. To mimic the human grasping process, we propose Directional Consistency optimization for Grasping (DCoG). Given an object point cloud and its corresponding contact map, DCoG guides the grasping process to simultaneously optimize both hand direction and grasp distance through well-designed energy terms.

    \subsection{Optimization Method}
    \label{sec:optim}

    The DCoG consists of multiple energy terms to achieve deterministic grasping results.
    
    \paragraph*{Contact-based Distance Energy} One key objective of grasping is to ensure the actual contact between the hand and the object. To achieve this, we propose a contact-based distance energy $E_{\text{dis}}$. This energy function aims to guide the hand to a suitable position by minimizing the squared Euclidean distance between the fingertips' points and the object contact map points, considering they belong to the same category. Mathematically, this energy can be expressed as:
    \begin{equation}
    E_{\text{dis}}=\sum_{c}^{C} \sum_{i}^{\left|\TT^{c}\right|} \text{Drop}\left(\min _{j}\left\|\TT_{i}^{c}-\OO_{j}^{c}\right\|_{2}^{2}, p\right)
    \label{Edis}
    \end{equation}
    where $\TT^{c}$ and $\OO^{c}$ denote the subsets of the point cloud associated with category $c$ within $\TT$ and $\OO$, respectively. $\TT^{c}$ is manually delineated, while $\OO^{c}$ is generated by the CVAE model mentioned in~\cref{sec:cvae}. To reduce the risk of getting trapped in a local optima, we employ the $\text{Drop}(\cdot,p)$ operation. It increases the randomness of the optimization process by introducing a probability parameter, denoted as $p$, to reset certain calculation results to zero.
    \begin{figure*}[h!]
\centering
    \includegraphics[width=0.98\textwidth,height=0.38\textwidth]{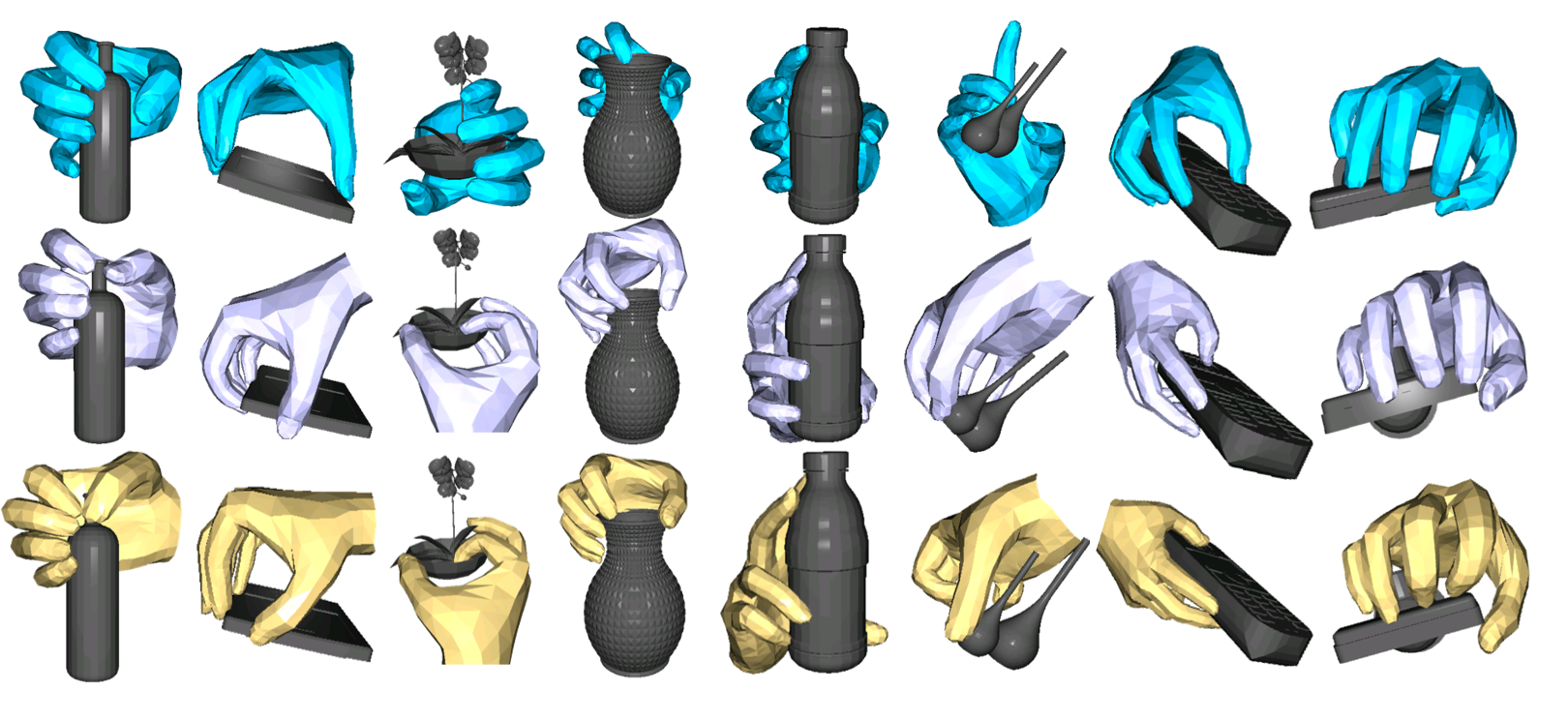}
    \vspace{-0.3cm}
    \caption{Qualitative experimental results. The \textit{top} shows grasps generated by our complete method, the \textit{middle} displays grasps obtained with the optimization-only method, and the \textit{bottom} represents the Ground Truth. }
    \label{fig:qualitative}
\end{figure*}

    \paragraph*{Directional Consistency Energy} In order to enable precise force applied to fingers during grasping, we introduce two energy terms to ensure the directional consistency: $E_\text{dct}$, defined over the fingertip, and $E_\text{dcf}$, defined over the finger. $E_\text{dct}$ and $E_\text{dcf}$ are defined as the squared Euclidean distance between the unit surface normal vector of a point in the set $\TT^{c}(\FF^{c})$ and the unit directional vector pointing towards the nearest point within the set $\OO^{c}(\OO)$. $\FF^{c}$ denote the subsets of the point cloud associated with category $c$ within $\FF$, which is also manually delineated, similar to $\TT^{c}$. Since the mesh of the hand can be obtained directly, the surface normal vectors of all points in $\TT$ and $\FF$ can be simply calculated. Let $N(x)$ denote the unit surface normal vector at point $x$, and let $u\left(x\right)= \frac{x}{\left\|x\right\|_{2}}$ indicate the unit normalization of $x$. Thus, $E_\text{dct}$, $E_\text{dcf}$ and the final directional consistency energy $E_\text{dc}$ are defined as
    \begin{equation}
    E_{\text{dct}}=\sum_{c}^{C} \sum_{i}^{\left|\TT^{c}\right|} {\text{Drop}}\left(\left\|N\left(\TT_{i}^{c}\right)-u\left(\OO_{k}^{c}-\TT_{i}^{c}\right)\right\|_{2}^{2}, p\right)
    \end{equation}
    \begin{equation}
    E_{\text{dcf}}=\sum_{c}^{C} \sum_{i}^{\left|\FF^{c}\right|} {\text{Drop}}\left(\left\|N\left(\FF_{i}^{c}\right)-u\left(\OO_{\dot{k}}-\FF_{i}^{c}\right)\right\|_{2}^{2}, p\right)
    \end{equation}
    \begin{equation}
    E_{\text{dc}}=\lambda_{\text{dct}}E_{\text{dct}}+\lambda_{\text{dcf}}E_{\text{dcf}}
    \label{Edc}
    \end{equation}
    where $k$ and $\dot{k}$ can be calculated via $k=\argmin_j\left\|\TT_{i}^{c}-\OO_{j}^{c}\right\|$ and $\dot{k}=\argmin_j\left\|\FF_{i}^{c}-\OO_{j}\right\|$, respectively. 

    \paragraph*{Adaptive Weight Adjustment Strategy} A potential challenge in the optimization process is determining the weight coordination of the two energy terms in~\cref{Edis} and~\cref{Edc}. To tackle this task, we propose an adaptive weight adjustment strategy. In the initial optimization phase, we assign a lower weight to $E_{\text{dis}}$ and a higher weight to $E_{\text{dc}}$, guiding the optimization algorithm to enhance the hand's grasping orientation. As the number of iterations increases, the weight of $E_{\text{dis}}$ progressively decreases, while the weight of $E_{\text{dc}}$ increases. This adaptive weight adjustment promotes a more natural grasping process.

    The non-convex nature of the dexterous grasping task presents challenges for the optimization algorithm in achieving a global optimal solution. Inspired by~\cite{mousavian2019graspnet}, we train a binary classification network based on the PointNet architecture which utilizes a binary cross-entropy loss function, taking the $\HH$ and $\OO$ as inputs, while predicting the success outcome of grasping. To avoid mismatching, we also incorporate the binary cross-entropy loss function as an energy term, denoted as $E_\text{net} = -\log {p_\text{net}}$, for network supervision in the optimization process. Here, $p_\text{net}$ represents the probability of being successfully grasped according to the network prediction. Since the network has learned the underlying principles of grasping, it gains an advantage in the optimization algorithm by effectively avoiding local optimal solutions. Another energy term considered is the penetration energy. As our method relies solely on point clouds, direct computation of the signed distance is not feasible. Consequently, we adopt $E_\text{pen}$ as proposed by Jiang \textit{et al}. \cite{jiang2021hand}.

    Finally, we formulate the total energy as:
    \begin{equation}
        E= i_{\text{c}}\lambda_{\text{dis}}E_{\text{dis}} +(i_\text{s}-i_\text{c})\lambda_{\text{dc}}E_{\text{dc}}+\lambda_{\text{net}}E_{\text{net}}+\lambda_{\text{pen}}E_{\text{pen}}
    \label{E}
    \end{equation}
    where $i_{\text{c}}$ denotes the current iteration index in the optimization process, and $i_{\text{s}}$ denotes the total number of iterations. These two parameters correspond to the adaptive weight adjustment strategy. 
   
    In this research, the parameters in~\cref{E} are set as:  $i_\text{s}=300,\lambda_{\text{dis}}=0.5,\lambda_{\text{dct}}=0.8,\lambda_{\text{dcf}}=0.6,\lambda_{\text{dc}}=1.0,\lambda_{\text{net}}=0.6,\lambda_{\text{pen}}=10$.

    \subsection{Initialization Strategy}

     Inspired by~\cite{wang2023dexgraspnet}, we adopt a similar hand pose preparing for grasping, as illustrated in Fig. 3(c). Then, we place the hand to a random distance from the contact map, aligning with the middle fingertip to ensure that it is away from the object. However, in real-world grasps, it is generally impractical to achieve a direct orientation of the palm facing the object. To address this, we introduce diversity by rotating the palm in various directions during initialization. The final grasping result is determined based on the minimum of $E_\text{pen}$, as incorrectly grasped results often exhibit a significant area of penetration with the object.

    \subsection{Analysis of Determinable Generation}

    Different grasping tasks require distinct finger contributions. Our method leverages individualized contact for each finger, employing highly interpretable energy terms. This enables us to achieve a customizable selection of individual finger contributions, which is in particular realized by adjusting the weight of each category when calculating $E_\text{dis}$. Such a strategy has proven to be effective in various scenarios, enabling the algorithm to determine which fingers should be given priority when grasping objects.
    
    
    

\section{Experiment}

    We conduct both quantitative and qualitative evaluations on the proposed approach, comparing it with other grasping methods. Additionally, we perform ablation experiments to illustrate the individual contributions of different modules and demonstrate the synergistic effect achieved through their combination.

    \subsection{Quantitative Evaluations}

    \paragraph*{Penetration} We evaluate both the penetration volume and maximum penetration depth for all test samples. A larger penetration indicates a deeper intrusion of the hand into the object. Conversely, if the penetration volume or penetration depth is 0, it may infer that the hand has no contact with the object.

    \paragraph*{Physical Displacement} We conduct a quantitative assessment of grasp stability using a physics simulator, following the methodology introduced in~\cite{jiang2021hand},\!\cite{tzionas2016capsimu}. In this evaluation, we place the hand and object within the simulator, then it calculates contact forces based on the penetration volume, and simulates the displacement of the object under these forces. A larger displacement suggests that the grasp is unable to apply sufficient forces to stably hold the object.

    \paragraph*{Success Rate} When a grasp involves a minimal penetration, it may result in a significant physical displacement if the hand cannot exert an appropriate grip force. Conversely, if a grasp exhibits considerable physical displacement, it may be due to excessive penetration between the hand and the object. Therefore, we define the success rate of a grasp by considering a trade-off between these two factors. In particular, a grasp is deemed successful if the penetration volume is less than 5\;$cm^3$ and the object displacement is smaller than 2\;$cm$. Otherwise, we categorize the grasp as a failure. We should admit that this definition of success rate may be stringent, emphasizing the granularity of the grasp.

    As mentioned earlier, to mitigate the impact of faulty grasps on our experimental results, we calculate the average of penetration and physical displacement across all successful grasps.
    
    \paragraph*{Perceptual Scores} We conduct a perceptual study to evaluate the naturalness of the generated grasps following~\cite{karunratanakul2020graspingfield} and~\cite{jiang2021hand}. In contrast to their approaches, we recruit 15 participants who are able to observe the complete view of the grasps in a 3D window, in contrast to only a few perspectives as that of~\cite{karunratanakul2020graspingfield} and~\cite{jiang2021hand}. Each participant is asked to score 20 randomly generated grasp results by different method shown in~\cref{table1}, where ContactOpt is not included due to low success rate, while the GA (w/o TTA) is excluded due to an obvious worse performance compared with GA (w/ TTA).

    The experimental results presented in~\cref{table1} exhibit that our success rate and perceptual scores surpass those of other methods. As ContactOpt requires complete MANO parameters as input, we utilize the generated parameters from our method for ContactOpt. Due to the potential instability of contact maps produced by the generation model, we conduct separate evaluations using the optimization-only method and the complete method. In the case of the optimization-only method, the contact maps are annotated using our annotation method presented in~\cref{sec:annotate}.

   We observe better results when using only the optimization method, primarily because the annotated contact maps obtained are more realistic. This highlights the effectiveness of our proposed optimization method. It also indicates that as the model’s capability to generate accurate contact maps improves, the proposed approach will be more likely to achieve better grasp results. 

\begin{figure}
    \centering
    \includegraphics[width=0.49\textwidth,height=0.56\linewidth]{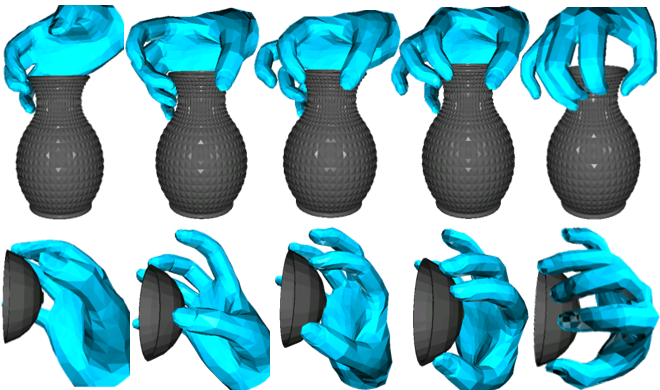}
    \caption{Determinable results. By sequentially setting the contribution of each finger to zero, we can obtain five determinable grasps.  }
    \label{fig:control}
\end{figure}
\subsection{Qualitative Evaluations}

    In~\cref{fig:qualitative}, we present visuals of grasps generated by our proposed complete method, the optimization-only method, and the Ground Truth (GT). The optimization-only method utilizes GT-annotated contact maps, resulting in generated grasps that closely resemble the GT. In contrast, the complete method autonomously generates contact maps using the CVAE, leading to differences between the generated grasps and the GT. Through qualitative experimental results, we demonstrate that the proposed method is capable of producing stable grasps. In addition, the determinable results are shown in~\cref{fig:control}, highlighting the deterministic capabilities of our method in generating grasps. By adjusting the contributions of different fingers, we generate grasp poses with determinable outcomes. We believe that this controllability in grasping will enhance the adaptability of the grasp algorithm, thereby enabling it to emulate the versatility and adaptability observed in human hand grasping.

\begin{table}[tb]
    \centering
    \caption{Ablation Study -  Energy Terms}
    \label{table2}
    \small
    \begin{tabular}{lccc}
        \toprule
        \;\textbf{Energy}& \textbf{Volume$(cm^3)$} & \textbf{disp.$(cm)\downarrow$} & \textbf{Succ. R.$(\%)\uparrow$} \\ \midrule
        w/o $E_{\text{dct}}$    & 2.10  & 0.89 & 40.80 \\  
        w/o $E_{\text{dcf}}$     & 2.55 & 0.85 & 39.67 \\
        w/o $E_{\text{dc}}$     & - & - & 7.34 \\
        w/o $E_{\text{net}}$    & \textbf{1.48}  & 0.93 & 43.52 \\   \midrule
        \!complete $E$             & \textbf{1.48} &  \textbf{0.82} & \textbf{45.51} \\   \bottomrule
        
    \end{tabular}
\end{table}    
	
\subsection{Ablation Study}

    The ablation study aims to evaluate the proposed energy terms: $E_{\text{dct}}$, $E_{\text{dcf}}$, $E_{\text{dc}}$, and $E_{\text{net}}$.

    The results of the ablation experiments in~\cref{table2} indicate that each energy term plays a distinct role. The removal of $E_{\text{dc}}$ results in a significantly reduced number of successful grasps, making other evaluation metrics meaningless in its absence. $E_{\text{dct}}$ and $E_{\text{dcf}}$ primarily contribute to increasing the success rate of grasps, while $E_{\text{net}}$, learned from the dataset, emphasizes the close interaction between the hand and the object, thereby enhancing the overall quality of grasps.

\subsection{Methodological Analysis}
    
    \paragraph*{Advantages} Our experimental results demonstrate the enhanced grasping granularity achieved by the proposed method. Notably, the proposed approach effectively preserves small object displacements in the simulation environment while minimizing penetration volume. Moreover, the proposed method only requires sampled point clouds of objects as input. In addition, our method exhibits a high degree of hand control, facilitating easy adjustment of each finger's contribution during the process of object grasping. This remarkable flexibility not only highlights the adaptability of our grasping method but also  establishes a robust groundwork for future extensions and enhancements.

    \paragraph*{Limitations} We also investigate the scalability of our proposed method. We observe that if the CVAE has not been trained on objects that have similar shapes to the input object, it may face challenges in generating accurate contact maps, leading to grasp failures. This indicates the dependency of our method on the generation capability of the CVAE model. To effectively overcome this limitation, comprehensive training on a large-scale dataset becomes necessary.
  
\section{Conclusion}
	
    In this study, we first revealed the limitations of existing methods in grasping generation. In order to tackle these challenges, we proposed a new scheme that utilized a CVAE model to predict individual contact maps for each finger. We also developed a new optimization method, DCoG, to guide the generation of grasps. Our approach \textbf{GrainGrasp} demonstrated promising performance in terms of granularity in grasping. In particular, we can determine the contribution of each finger to the final grasp result, which was not presented in previous methods. We should point out that the performance of the proposed scheme relied on the generation capability of CVAE during the contact map generation process. We leave this for future research.

\bibliographystyle{IEEEtran}
\balance
\bibliography{reference}
\end{document}